\documentclass{article}

    \usepackage[nonatbib, preprint]{neurips_2019}

\usepackage[utf8]{inputenc} 
\usepackage[T1]{fontenc}    
\usepackage{hyperref}       
\usepackage{url}            
\usepackage{booktabs}       
\usepackage{amsfonts}       
\usepackage{nicefrac}       
\usepackage{microtype}      
\usepackage{amsmath}       

\usepackage{float}
\usepackage{graphicx}
\usepackage{subfig}

\usepackage{cite}

\title{Synthetic Embedding-based Data Generation Methods for Student Performance}

\author{
  \textbf{Dom Huh}\\ 
  $^{1}$Department of Electrical and Computer Engineering,\\ George Mason University, Virginia, USA \\
  \texttt{dhuh4@gmu.edu}
}

\begin{document}

\maketitle

\begin{abstract}
Given the inherent class imbalance issue within student performance datasets, samples belonging to the edges of the target class distribution pose a challenge for predictive machine learning algorithms to learn. In this paper, we introduce a general framework for synthetic embedding-based data generation (SEDG), a search-based approach to generate new synthetic samples using embeddings to correct the detriment effects of class imbalances optimally. We compare the SEDG framework to past synthetic data generation methods, including deep generative models, and traditional sampling methods. In our results, we find SEDG to outperform the traditional re-sampling methods for deep neural networks and perform competitively for common machine learning classifiers on the student performance task in several standard performance metrics.
\end{abstract}

\section{Introduction}\label{introduction}
In the educational domain, student performance tasks involve designing a classifier to predict the performance of a student given a set of features. These datasets, however, are often skewed centrally to the mean as a relative, more commonly known as bell-curve, grading scale is often employed. Consequently, the limited number of examples belonging to outlier students, specifically those demonstrating extreme success and failure, do not sufficiently represent minority classes, as opposed to the overwhelming number of average performing students' examples that represent to the majority classes. This evident disproportion results in machine learning algorithms failing to understand and learn how to classify the minority classes, which was supported in \cite{DBLP:journals/corr/abs-1106-4557}. This issue we are describing is known in the machine learning community as class imbalances. Class imbalances are recognized when working with an imbalanced dataset, that exhibits a notable disparity in the number of examples amongst different classes, and often can disproportionately harm the classifier's performance on the minority classes. As \cite{review} discusses, these datasets describe a nature of the problem, which correlates heavily with rare events, small sample size, low class separability, and/or existence of within-class subconcepts. For student performance datasets, we find all to be the case, and even so, gaining insight into the outlier students is significant for the educators as it may point to indicators of failure and success.

\begin{figure}
    \centering
    \includegraphics[scale=0.6]{"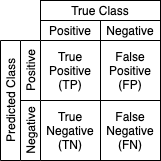"}
    \caption{Confusion Matrix: A contingency table between true class and predicted class in a binary class setting.}
    \label{fig:confusionmatrix}
\end{figure}

Commonly for class imbalance tasks, the nature of the dataset results in a relatively high predictive accuracy for the majority classes and significantly lower in the minority classes. If classification accuracy was used to evaluate the model's efficiency, we would obtain an overly optimistic belief in the classifier. Thus, often for such datasets \cite{10.5555/3001392.3001400}, we consider using Area under Curve (AUC) on the Receiver Operating Characteristic (ROC) curve, a graph used to show the trade-off between true positive and false positive error, as the performance metric for the classifier. The ROC space requires the true positive rate (TPR), or more commonly known as the sensitivity, and the false positive rate (FPR), or more commonly known as fall-out. We can calculate the values for TPR and FPR, seen in Equation \ref{roc}, from the components in confusion matrix, a table containing the values of true and false positives and negatives, seen in Figure \ref{fig:confusionmatrix}.

\begin{equation}\label{roc}
\begin{split}
    TPR = \frac{TP}{TP+FN}\\
    FPR = \frac{FP}{FP+TN}
\end{split}
\end{equation}

\begin{figure}
    \centering
    \includegraphics[scale=0.6]{"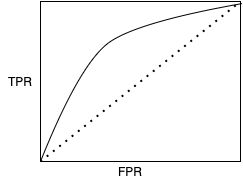"}
    \caption{ROC curve: Above shows a graph in ROC space. The dotted line denotes the ROC curve if the classification is randomly estimated. The solid line denotes the ROC curve if the classification is more accurately estimated. The AUC would be the area under the ROC curves.}
    \label{fig:roc}
\end{figure}

Then, we obtain the ROC curve to calculate the AUC score, seen in Figure \ref{fig:roc}. For multi-class settings, we can use an one-vs-rest paradigm, and either obtain the AUC scores at the class level, or aggregate with or without class weighing to obtain a micro-AUC or macro-AUC score respectively. When we view at a class level, we call this class-AUC score.

Generally, as defined in \cite{review}, there exists three common approaches to alleviate the effects of class imbalances: data modification, algorithm modifications, learning modification. As the name suggests, each approach modifies different aspects of the learning system to counter class imbalances. In this paper, we will focus on data modification approaches, or more specifically sampling methods.

We consider two classes of sampling methods: oversampling and under-sampling. Oversampling methods append samples to the pre-existing training set, whereas under-sampling methods remove samples from the training set. Both approaches aim to balance the number of samples, given some criteria, in the training set. More generally, sampling methods must consider two important components. Firstly, the imbalanced criteria that is targeted must be defined. For imbalanced dataset, for example, the criteria may be the number of samples in the class. Secondly, the method of creating the balance must then be defined, such as randomly re-sampling or randomly removing samples. Variations of these two components have been explored in the past works \cite{https://doi.org/10.1002/eng2.12298}. In particular, a promising oversampling approach that has improved handling of imbalanced datasets has been generating new synthetic samples from pre-existing samples from the training dataset, and this method is known as Synthetic Data Generation (SDG). In \cite{DBLP:journals/corr/abs-1106-1813}, the authors introduced the prevalently used Synthetic Minority Over-sampling Technique (SMOTE), which generates new samples by modifying a pre-existing sample by adding the scaled (0-1) difference between feature vectors of its nearest neighbors, with the addition of noise. Furthermore, \cite{10.1145/1007730.1007736, 4633969} proposed variants of SMOTE, called DataBoost-IM and ADASYN respectively, changing the sampling criteria to be based on predictive error.

Recently, deep learning methods have exhibited advances in robustness \cite{DBLP:journals/corr/abs-1903-09730, shamsolmoali2020imbalanced} and efficacy, specifically to end-to-end training approaches \cite{8566353}, for SDG methods. In specific, deep generative models (DGMs), such as generative adversarial models, have been previously introduced to tackle class imbalances \cite{DBLP:journals/corr/abs-1903-09730, shamsolmoali2020imbalanced}.

In this paper, we introduce a framework for embedding-based SDG methods, called SEDG, to handle effects of class imbalances on student performance tasks. The student performance dataset and the set of classifiers used will be described in Section \ref{Preliminaries}. Further details of SEDG framework and deep generative models will be discussed in Section \ref{SDG}. In Section \ref{Results}, the results are presented and evaluated.

\section{Preliminaries}\label{Preliminaries}
In this section, we will discuss the details of the Student Performance Dataset from \cite{dataset} and the classifiers used to test the SEDG framework introduced in this paper. The training methods and implementation decisions will also be described.

\begin{figure}
    \centering
    \includegraphics[scale=0.6]{"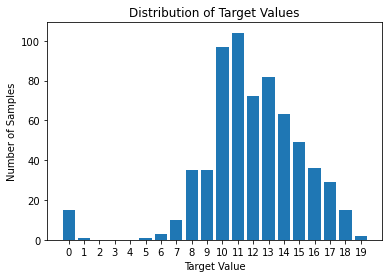"}
    \caption{Target Value Distribution: From the student performance dataset, the imbalance in the distribution of target values skews centrally, indicating a disproportionate number of examples between the central classes and the outer classes.}
    \label{fig:targetdistrbution}
\end{figure}

\subsection{Student Performance Dataset}
\begin{figure}
    \centering
    \includegraphics[scale=0.4, width=0.8\textwidth]{"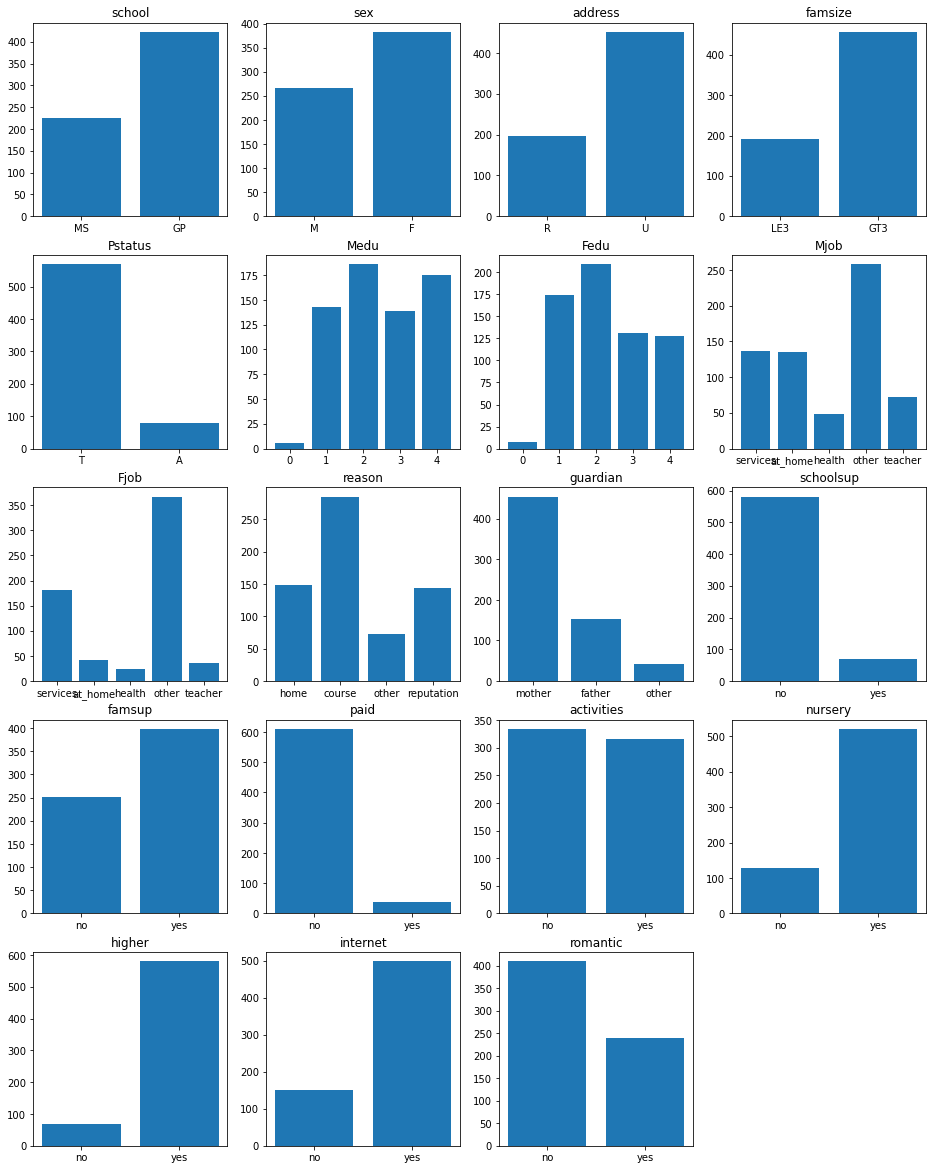"}
    \caption{Feature Distribution: From the student performance dataset, we can see an imbalance in some of the distribution of feature values.}
    \label{fig:featuredistrbution}
\end{figure}

The dataset provided by \cite{dataset}, can be used to predict student performance, or the final grade, on a scale from $0$ to $20$. The data was collected from two Portuguese secondary level schools, and has $649$ samples of $32$ features. In \ref{fig:targetdistrbution}, we show the class imbalance in the dataset, skewed centrally as mentioned in Section \ref{introduction}. In \ref{fig:featuredistrbution}, we show the distribution of features in the dataset, where some features, such as parent’s cohabitation status, extra paid classes, and extra educational school support, also demonstrated to be imbalanced. Further information on the meaning of these features to can be found in \cite{dataset}. We will not opt to use the binary or five-level  classification found in other works on this dataset, but will classify all 20 classes. For future works, we find a three-level classification, or binary classification between the outlier samples, which exists on the edges of the target values, and the average samples to be more appropriate simplification.

\subsection{Classifiers}\label{classifiers}
We will use three traditional machine learning classifiers (support vector machines, random forest, gradient-boosted decision trees/XGBoost) and an neural network architecture which we will denote as NNModel.

Support vector machines (SVM) \cite{10.1109/5254.708428} are optimal-margin classifiers that seek to find the hyper-plane that maximize the geometric margins between binary classes. In a multi-class setting, we can proceed by using an one-vs-rest approach, where the maximum score from $n$ SVMs, where each SVM is trained on masking all but one class to formulate a binary setting, is selected, or an one-vs-one approach, where similar to one-vs rest, we formulate a binary setting, but with pairs of classes instead.

Decision trees \cite{10.1023/A:1022643204877} are tree-based classifier that create binary splits based on a condition on the features that maximizes information gain, a metric of weighted entropy. Random forest \cite{10.1023/A:1010933404324} is a bootstrap aggregation, or bagging, algorithm using decision trees, where $n$ learners learn on bootstrapped subsets of the dataset and are also limited to a subset of features it can split on. Thus, inference of random forest is run on majority vote of the learners. XGBoost \cite{DBLP:journals/corr/ChenG16} is a gradient boosting algorithm using decision trees, where learners are introduced to additively correct errors of trained learners. Gradient boosting refers to correcting the errors through gradient learning when adding new learners \cite{10.2307/2699986}.

\begin{figure}
    \centering
    \includegraphics[width=0.6\textwidth]{"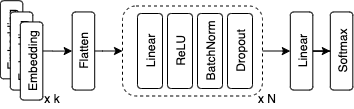"}
    \caption{NNModel Block Diagram: In the diagram, the input is composed of $k$ features. The input data is first mapped into embeddings, then passed into the processing component, which has $N$ LRND blocks, denoted as the dotted block. To obtain the final prediction for a multi-class classification, the output is transformed using a softmax activation.}
    \label{fig:nnmodel}
\end{figure}

For our experiments, the neural network architecture we call NNModel is a multi-layer perceptron model that is sequentially made up of $N$ LRND blocks, seen in Figure \ref{fig:nnmodel}. The LRND block is composed of a linear layer, rectified linear activation, batch normalization, and followed by a dropout layer. This model is used for classification, and its parameters are optimized using gradient learning on cross entropy loss, seen in Equation \ref{ce}, where $p$ is the true, target function and $q$ is the hypothesis function, and an Adam optimizer scheduled to reduce and anneal the learning rate on plateaus by an order magnitude.

\begin{equation}
    L(p,q) = -\frac{1}{N}\sum_{\forall x} p(x) \log(q(x)), |x| = N
    \label{ce}
\end{equation}

\section{Synthetic Embedding-based Data Generation} \label{SDG}
In this section, we will discuss the embedding-based method for SDG that will be evaluated in Section \ref{Results}. We will formalize the generalized SDG method step by step, dividing the discussion into 4 chronological parts: sample selection, feature selection, feature modification and synthetic sample usage. In each step, we offer various design considerations. We will then write on deep generative models in the context of SDG.

\subsection{Sample Selection}\label{ss}
We must first consider how to select the samples from the original dataset to base the synthetic samples on, and also, how many samples we want to select. The latter consideration of the number of samples can be fixed or stochastically chosen within a defined range. we will call the former consideration \textit{sample selection} henceforth.

Naively, we can randomly select $k$ samples from the entire dataset $D$, all with equal probability. Once the samples are selected, we place them into a sample pool, $S$, that will be used to generate the new synthetic samples. This approach will be referred to as random sample selection (RSS).

A more systematic approach is to partition the dataset $D$ in a targeted manner and select a member $M_i$ of the partition $P = \{M_0,M_1...\}$ to sample from. The member selection can occur at a probability $p$, or deterministically, which can then be stochastically sample from, with or without replacement. More concretely, suppose we want $k$ samples and given we are working with imbalanced datasets, we can choose the partition $P$ to be divided based on the classes. We set each member of the partition $M_i\in P$ to be associated to a probability $p_i$ proportional to their cardinality $p_i \propto |M_i|$, which represent how likely the member will be selected. Thus, there exists a surjective mapping from the partition$P = \{M_0,M_1...M_n\}$ to the probability distribution $\bar{p} = \{p_0,p_1...p_m\}$. A special case would be to deterministically select the member $M_i$ based on the cardinality, where $|M_i| \ge |M_j|$ for all $j\in [0,n]$. Once a member $M_i$ is selected, we sample $m$ examples from the set, where $m\leq k$. The value of $m$, similar to $k$, can be fixed or randomly chosen within a defined range. The $m$ samples are stochastically selected from $M$ by performing RSS on the subset of $D$, we add this to our sample pool $S$. We repeat our selection of a new member in the partition, with or without replacement, where at the end of the process, we get $|S| = k$. This approach will be referred to as partition-based sample selection (PaSS). We note that this approach is dependent solely on the dataset, and remains entirely independent on the classifier.

Another approach for sample selection consists of selecting samples based on the classifier's performance, similar to concept of boosting. Specifically, we can associate higher selection likelihoods for samples the learner has the largest margin for error, or uncertainty. In this context, we can consider classification accuracy to quantify error and uncertainty. Thus, we can place the highest probability of selection to the samples that were misclassified, and rank the correctly classified samples based on the classifier's certainty of the prediction. So, given a classifier $C$ and an loss function $L(C,s)$, where $s\in S$ is a sample belonging to the training set $S$, we let the selection probability $p_i$ for $x_i$ to be greater than the selection probability $p_j$ for $x_j$ if and only if $L(C,x_i)>L(C,x_j)$. The loss function returns a scalar value that represents the correlation between the target and prediction values. This approach will be referred to as performance-based sample selection (PeSS). We note that this approach, in contrast to PaSS, is dependent on the classifier, thus also requires trained model for its operation. Consequently, PeSS is more computationally costly to use.

We now can formulate a selection method that incorporates from both PaSS and PeSS approaches by using class-AUC score. We can first use class partitions described above and the class-AUC score as our uncertainty metric to develop a selection probability distribution to select the member to sample from. Thus, we incorporate elements from the PaSS approach by using a class partition to first select a member to sample from, and we incorporate the AUC score from the classifier to develop the probability associated to member selection. We call this the partition-performance sample selection approach, which we will refer to as PPSS.

\subsection{Feature Selection}\label{fs}
We can think of every sample $S \in \bar{S}$ from the dataset $D$, where $\bar{S} \subseteq D$, at a more granular level, to be made up of features. Or in other words, a sample $S = \{f_0,f_1,f_2...\}$ can be defined to be a set of features. After sample selection, we can modify the selected samples to generate new samples. Thus, given some mapping $\phi$ from $S$ to $S'$ where $\phi: \{f_0,f_1,f_2...\} \rightarrow \{f_0^{'},f_1^{'},f_2^{'}...\}$ would provide us new samples $S^{'}$, where $S^{'} \not = S$. But in order to modify these features, we now must consider how to select which features to modify efficiently as to appropriately increase variance in the dataset without creating class overlaps. But, to modify the samples, we must first select the features of the sample that we wish to adjust. We will discuss three approaches to select features: random selection, feature imbalance, and feature importance approach. We can group the feature imbalance and feature importance together, and label this weighted feature selection. In all, we will call this procedure \textit{feature selection} henceforth.

The random selection approach, similar to RSS, stochastically selects a subset of feature $\bar{f} \subseteq S = \{f_0,f_1,f_2...\}$, where $|\bar{f}| \leq |S| = n$, with all features with equal probability. The number of feature selected $ k = |\bar{f}|$ can be fixed or randomly chosen per sample.

The feature imbalance approach is to select the feature based on feature value imbalances, as seen in Figure \ref{fig:featuredistrbution}. We can quantify the imbalances by using the following method, which differs from past works \cite{review}. We first create $x = \{x_0, x_1...\}$, where $|x|$ is the number of features, by counting the number of examples that contains each unique value, resulting in a set of counts $\{c_0,c_1...\}$, for each feature. Then, for each $x_i \in x$, we find the clusters for $x_i$ using  k-means, with $k=2$, to obtain the ratio $r_i$ between the difference between the two centroids, and the maximum count in $x_i$. In the end, we obtain a vector of ratios $R = \{r_0,r_1...\}$, where $|R| = |x|$, which we can first normalize to treat as a probability distribution to select subsets of features. This approach is largely motivated to place higher emphasis on more imbalanced features, and thus attempting to balance all features' value representation.

The feature importance approach creates a probability distribution $\bar{p}$ based on feature importance, which will be used to select the features. Feature importance provides insight into the relationship between the predictive system and the data, and how much significance each feature in the data may have on the overall system's performance at the given task at hand. We will consider three approaches to calculate and obtain the feature importance: mean decrease impurity importance, permutation importance, drop-column importance.

\begin{figure}[]
    \centering
    \includegraphics[width = 0.9\textwidth]{"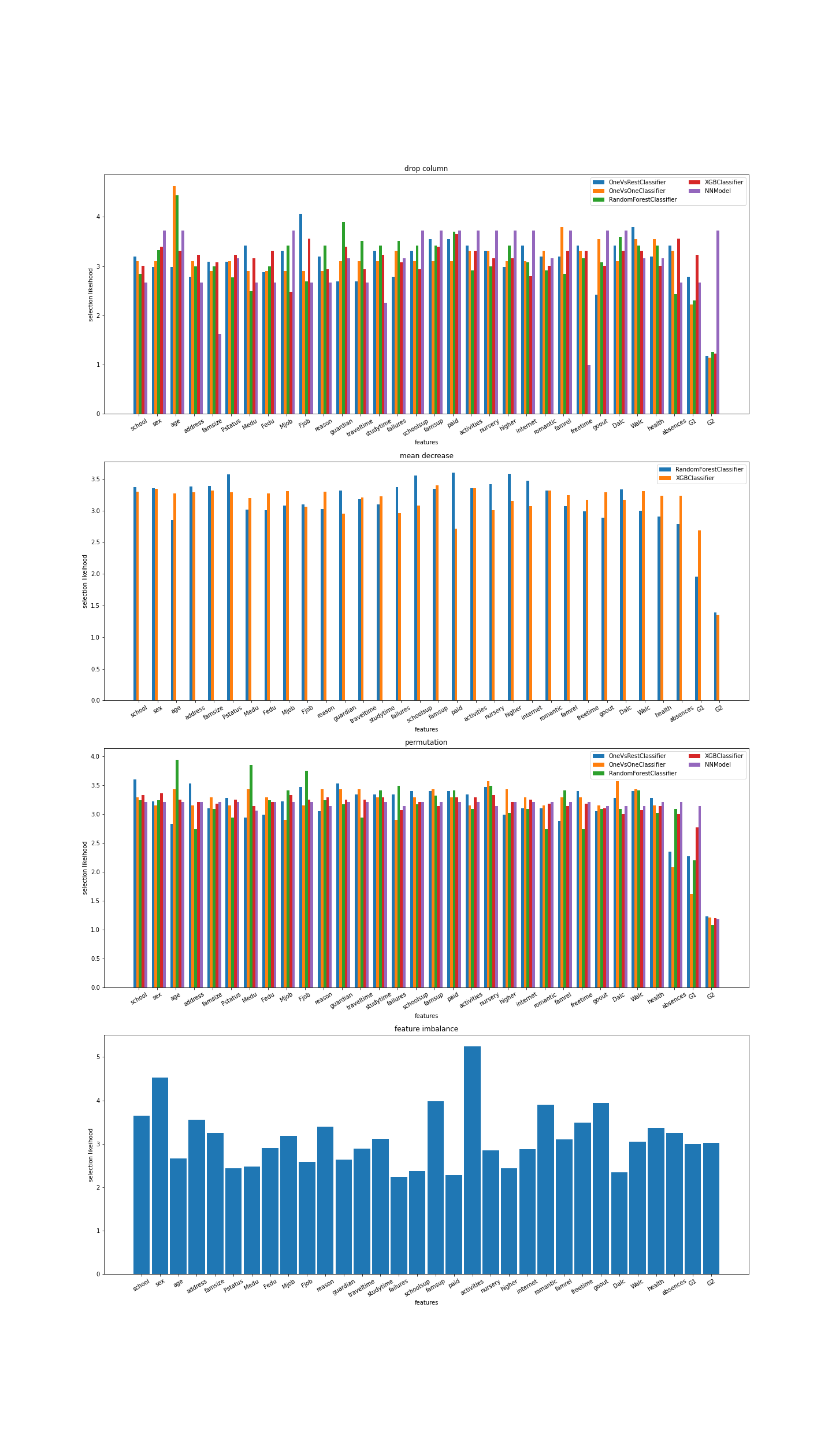"}
    \caption{Selection distribution from weighted feature selection methods for classifiers.}
    \label{fig:improvement}
\end{figure}

\subsubsection{Feature Importance}\label{fi}
Mean decrease, or gini impurity, importance methods rely on the use of tree-based classifiers, and is calculated by aggregating the gini decreases of each feature, which is criteria used to determine the splits, at every tree. However, \cite{bias_fi} has shown gini impurity importance approach is biased given the scale of measurement or number of categories of the features. This approach can only be used for tree-based classifiers, thus will only be consider such models.

Permutation importance methods, on the other hand, calculate the feature importance by evaluating the decrease in performance of a trained model on a test set that is shuffled on at the single feature value of interest. The shuffling process can be repeated to test multiple permutations of the feature values. Thus, permutation importance only requires the model to be trained once, but also needs to use the testing dataset. Also, this approach can be used for any models. 

Similarly, drop-column importance methods obtain the feature importance using the difference between the baseline performance of a model, which would trained on the entire dataset, and the performance of the model that has been trained on a limited dataset with a single feature value dropped. Thus, drop-column importance can also be used for any model. However, we see that drop-column importance is very computationally costly, proportionally to the number of features in the dataset.

\subsection{Feature Modification}\label{fm}
We must now consider how to modify the selected features optimally, a procedure we will call \textit{feature modification} henceforth. We define optimality as maximizing the variance between the synthetic samples from original samples and minimizing the class overlap in the new dataset. Our aim is to maximize the improvements in the minority classes, since that is the deficient area. To modify the features, we can naively inject noise to the continuous features and replace discrete features with sampling, but we wish to formulate a more targeted approach. In this paper, we will focus on embedding-based modifications, a search-based method where we will leverage some learned embeddings to offer insights in how to optimally modify the features.

Embeddings are mappings from the feature domain to a domain that can be more useful and understandable for the classifier to perform the task at hand. In other words, we can think of embeddings as optimized data pre-processing transforms. We consider two methods of embedding generation: classification transfer learning and auto-encoding. However, there are many other embedding generation approaches \cite{DBLP:journals/corr/abs-1206-5538} that have been successfully demonstrated in field of representation learning.

\subsubsection{Embedding Generation}\label{embgen}
Classification transfer learning refers to learning the embedding mapping $\phi_{\theta_\phi}$ through training an classifier $F_{\theta_F}$ on the task at hand. We note that both $\phi_{\theta_\phi}$ and $F_{\theta_F}$are parameters by their own independent weights $\theta_{\phi}$ and $\theta_{F}$. Thus, inference on the model can be seen in Equation \ref{inference}, where $\hat{y}$ is the prediction, and $x$ is the input data.

\begin{equation}\label{inference}
    \hat{y} = F_{\theta_F} (\phi_{\theta_\phi} (x))
\end{equation}

Thus, the classifier and the embedding mapping are both optimized for the targeted tasks, given some loss function $L$. Thus, if gradient learning is used, the weight updates can be seen in Equation \ref{update}, where $L$ is the loss function for classification, such as cross entropy, which is seen in Equation \ref{ce}.

\begin{equation}\label{update}
\begin{split}
    \theta_\phi = \theta_\phi - \nabla_{\theta_\phi} L(y,\hat{y}) \\
    \theta_F = \theta_F - \nabla_{\theta_F} L(y,\hat{y})
\end{split}
\end{equation}

Once the classifier and embedding mapping are trained, we can export the embedding mapping to be used for feature modification.

Auto-encoders $A_{\theta_A}$ are a type of neural network that learn to compress and decompress the input data, thus having a bottleneck structure. In other words, let the set of layers $l = \{l_0,l_1...l_n\}$ be the layer that make up the autoencoder, and $|l_i|$ for $i \in (0,n)$ be the number of parameters in layer $l_i$, then, in our definition, given $0\leq a<c\leq n$ and $b \in (a,c)$, there must exists $L_b$ such that $|l_b|<|l_a|$ and $|l_b|<|l_c|$, where we will refer to $l_b$ as the bottleneck layer. The aim of the autoencoder is to accurately build a reconstruction of the input given this decompression. Inference on auto-encoders can be seen in Equation \ref{autoencoding}, where $\hat{x}$ is the prediction, and $x$ is the input data. We note that the dimension of $\hat{x}$ and $x$ are equal.

\begin{equation}\label{autoencoding}
    \hat{x} = A_{\theta_A} (\phi_{\theta_\phi} (x))
\end{equation}

Auto-encoders are optimized given some loss function $L$, often referred to as the reconstruction loss, and the target is the input data $x$. Typically, the loss function $L$ will compare the input $x$ to the predicted value $\hat{x}$ directly, using a function like mean-square error or cross entropy, again seen in Equation \ref{ce}. Thus, if gradient learning is used, the weight updates can be seen in Equation \ref{ae_update}.

\begin{equation}\label{ae_update}
    \theta_A = \theta_A - \nabla_{\theta_A} L(x,\hat{x})
\end{equation}

Once the auto-encoder is optimized, we can export a proper subset of layers $E = \{l_0, l_1,...l_b\}$, where $l_b$ is the bottleneck layer, as the embedding mapping.

We will also consider variational auto-encoders (VAE), which follows the same inference and optimization methods, however, we represent the bottleneck layer $l_b$ as a probability distribution, often a Gaussian $N(\mu,\sigma)$. Thus, the input of layer $l_{b+1}$ will be samples from the distribution from $l_b$. The loss function often used is empirical lower bound (ELBO), which combines the reconstruction loss normally used in auto-encoders and the Kullback–Leibler divergence loss term $KL(P_E(z|x),P_D(z))$, where $E$ is a mapping that uses the layers $\{l_0,l_1,...l_b\}$, $D$ is a mapping that uses the layers $\{l_{b+1},l_{b+2},...l_n\}$ and $z$ is the latent space representation, or the output of the $E$. Thus, $z$ will be considered to be the embedding of the sample.

When we use the auto-encoders are the embedding function, we remark that the models are trained on the training set independently to the training of the classifier. Thus, for computational efficiency, we can store and reuse embedding models.

For each of the embedding generation approaches, we can use principal component analysis for dimensionality reduction and further compress the embedding.

Returning back to our discussion on feature modification, as mentioned before, a sample $S$ can be thought of as a set of features $\bar{f}_s$. These features can either be discrete and continuous. We will consider both cases, and discuss how we can handle each case separately.

Assuming a set of features $\bar{f_d} \subseteq S$ that was selected for modification is homogeneously discrete, for each element $f_{d,i}$ in $\bar{f_d}$, there exists a set of possible values it can take on $y_i = \{y_{i,0},y_{i,1},...y_{i,k}\}$. We can randomly select a subset $c_i$ of $y_i$, where $|c_i| \leq |y_i|$ and if $f_{d,i} = y_{i,m}$ then $y_{i,m} \not \in c_i$. If the search space for unique feature values $y_i$ is small enough, we can try all possible values for $f_{d,i}$, however if it is large, then we could sample a subset as mentioned. Given the embedding mapping $\phi_i: y_i \rightarrow E_i$, we calculate the similarity score between $\phi_i(f_{d,i})$ and $\phi_i(y_{i,j})$ for all $y_{i,j} \in c_i$, and replace the feature value with the one with the highest similarity score. We repeat this process for every feature in $\bar{f_d}$. However, given the embedding mapping $\phi: S \rightarrow E$, where $\bar{f_d} \subseteq S$, we will again follow the same procedure, however we must compare $\phi(S)$ and $\phi(S')$ where $S = \{f_0,f_1...f_i,...\}$ and $S' = \{f_0,f_1...f_i',...\}$ where $f_i' \in c_i$. When we repeat this process for each feature in $\bar{f_d}$, we can choose to replace $S$ to the updated $S'$ for the proceeding feature modifications, or choose not to replace $S$.

Assuming a set of features $\bar{f_c} \subseteq S$ that was selected for modification is homogeneously continuous, for each element $f_{d,i}$ in $\bar{f_d}$, we define a range with a fixed step $y_i = \{y_{i,0},y_{i,1},...y_{i,k}\}$ where $f_{d,i} \in (y_{i,0}, y_{i,k})$ and $y_{i,0}<y_{i,k}$. Then, we proceed similarly to the discrete feature case for feature modification.

\subsection{Synthetic Sample Usage}\label{ssu}
Once we obtain the synthetic samples, we now must decide how the samples will be utilized. We propose two considerations: cold or warm start and iterative or non-iterative,.

Initially, the model learns on the training set $D_{train}$. Then, the set of synthetic samples $D_{s,t}$ are created and added to the training set $D_{train}$. We can choose to either re-initialize the weights of the model, which we will call cold-start, or keep the weights from the previous training cycle, which we will call warm-start. We must now decide whether to repeat the process of creating a new $D_{s,t+1}$ from the newly trained model, making the process iterative, or end the training cycle entirely, making the process non-iterative. If the process is iterative, we can also choose to dropout and replace the previous $D_{s,t}$ partially or wholly with $D_{s,t+1}$, or append $D_{s,t+1}$ to the current $D_{s,t}$ at each step. The latter option can quickly become memory intensive.

\subsection{Deep Generative Models}\label{dgm}
In this section, we will formalize and discuss deep generative models (DGM) in more depth. Generative models can either be defined as modeling the conditional probability $P(x|y)$, or the joint probability $P(x,y)$ or the prior $P(x)$. A DGM is expressed using a deep neural network, parameterized by learnable weights $\theta$. Equivalently, DGM are mappings $f_\theta: y \rightarrow x$, $f_\theta: x \rightarrow x$, or $f_\theta: x,y \rightarrow x$. In context of SDG, DGM can be viewed as an end to end approach to generating synthetic samples, bypassing the need for Section \ref{fs}, and \ref{fm}. For our purposes, the objective is to learn how to generate new samples following the definition of optimality discussed in Section \ref{fm}. Thus, we can set the input of the DGM to either be random noise, or samples selected using methods from Section \ref{ss}. A potential issue that is highly problematic is the possibility of the generative model learning to map the input to itself without any variance, especially for traditional auto-encoders and given the nature of the optimization. We suggest adding some regularization term to the loss function to more directly counter the over-fitting problem for future works, however this paper does and will not propose a proven solution to this issue. In fact, we address this issue by simply preventing over-fitting with early stopping with heuristics, discussed in further detail in Section \ref{Results}. Now, we will discuss two main approaches to training DGMs: unsupervised and adversarial training.

\subsubsection{Deep Generative Model Training}
Unsupervised training refers to learning the prior $P(x)$, thus in more practical terms, optimizing the model using only the input $x$. An example of unsupervised training is training auto-encoders, where we treat the input as the output as well. Instead of exporting the embedding function, as seen in Section \ref{embgen}, we treat the auto-encoder as the generator model $f_\theta: x \rightarrow \hat{x}$, where $\hat{x}$ is the synthetic sample. In this paper, we will solely focus on auto-encoders and VAEs as traditional unsupervised learning methods for deep generative modeling.

Adversarial training adds onto unsupervised learning with the discriminator model, which learns to discern real or fake data. At a high level, the generator model attempts to trick the discriminator model by learning how to generate samples that are realistic. We will consider the traditional approach and the conditional approach.

The traditional approach is to train the generator and discriminator models separately using unsupervised and supervised learning respectively. The training for the generator model will follow the procedure discussed in Section \ref{embgen}, however we aggregate the discriminator's inaccuracy and the reconstruction loss, scaled by $\alpha$ and $1-\alpha$ respectively, where $|\alpha| \leq 1$, to obtain our generative model's loss. The training of the discriminator will optimize normal classification between real and fake data using the dataset and the generator model. We can instead have the discriminator instead predict on the latent representation,or the embedding, as seen in \cite{8566353}. For this approach, we will focus on adversarial auto-encoders and adversarial VAEs.

The conditional approach follows most of what is stated in the traditional approach, however we aim to have both the generative and discriminative models conditioned on the class label, $y$. Similar to practices discussed in \cite{8566353}, we can simply treat $y$ as another input layer, paired with an embedding layer, and is trained accordingly. For this approach, we will focus on conditional generative adversarial networks using auto-encoders and VAEs as the generative model.

\section{Results}\label{Results}
In this section, we will evaluate the methods discussed in Section \ref{SDG} using the student performance dataset and classifiers described in Section \ref{Preliminaries}. We will compare our results with other balancing methods from past works, such as random oversampling, random under-sampling, SMOTE, Tomek Links, extended nearest neighbors, and various combinations methods of these approaches.

\begin{figure}
    \centering
    \includegraphics[width = \textwidth]{"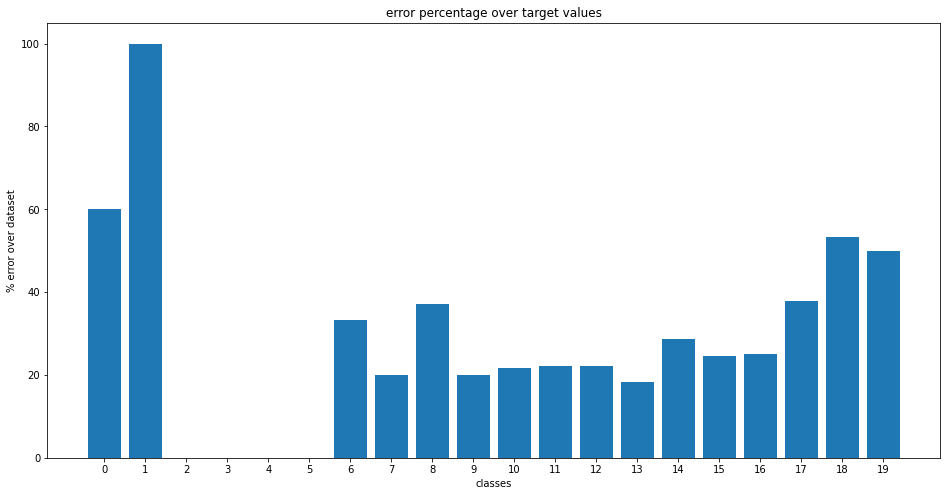"}
    \caption{Percent error distribution on the testing set over all target classes averaged over classifiers from Section \ref{classifiers}}
    \label{fig:error}
\end{figure}

We see that if the classifiers are trained on the dataset normally, optimizing the cross entropy loss, we obtain performance seen in Figure \ref{fig:error}, which follows our hypothesis of disproportion performance between minority and majority classes. We seek to mitigate these effects of class imbalances with the SDG methods we have proposed.

\begin{figure}
    \centering
    \includegraphics[width = \textwidth]{"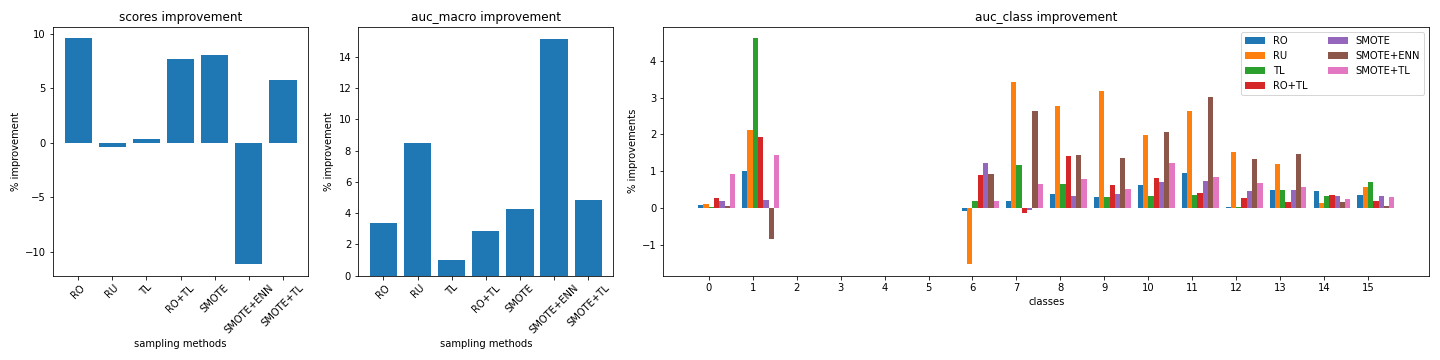"}
    \caption{Max \% performance improvement with traditional sampling methods for imbalanced student performance dataset on the testing set.}
    \label{fig:benchmark}
\end{figure}

We used the balancing methods aforementioned in Section \ref{introduction}, and we obtained the max \% improvements over 50 trials. The results can be seen in Figure \ref{fig:benchmark}. However, as we can see, they all provide a performance boost to the all performance metrics, except for SMOTE+ENN and random under-sampling, which detriments the classifier's performance. To avoid under-sampling methods to be completely useless as there exists minority classes with $\leq 10$ samples, thus under-sampling an already limited number of samples would lose immense information, we limited the under-sampling to majority classes.. 

We now will discuss our results from our experiments using the SEDG and DGM methods mentioned in Section \ref{SDG} and using the NNModel and the traditional models mentioned in Section \ref{classifiers} as the classifier. For all of the experiments below, we set the number of synthetic samples that will be generated to $200$ samples. All recorded improvements are on the testing set, which encompasses $40\%$ of the dataset. We define $\%$ improvement $PI$ in Equation \ref{metric}, where $M(y_{syn},\hat{y}_{syn})$ is the score from some performance metric $M$ from the dataset with the synthetic samples, and $M(y,\hat{y})$ is the score from the performance metric $M$ from the original dataset.

\begin{equation}
    PI = M(y_{syn},\hat{y}_{syn}) - M(y,\hat{y})
    \label{metric}
\end{equation}

\subsection{NNModel as Classifier}\label{nntest}
Now, we consider using the NNModel defined in Section \ref{classifiers} as our classifier, and we will test our embedding-based SDG methods and DGMs on the student performance dataset.

\begin{figure}
    \centering
    \includegraphics[width =\textwidth]{"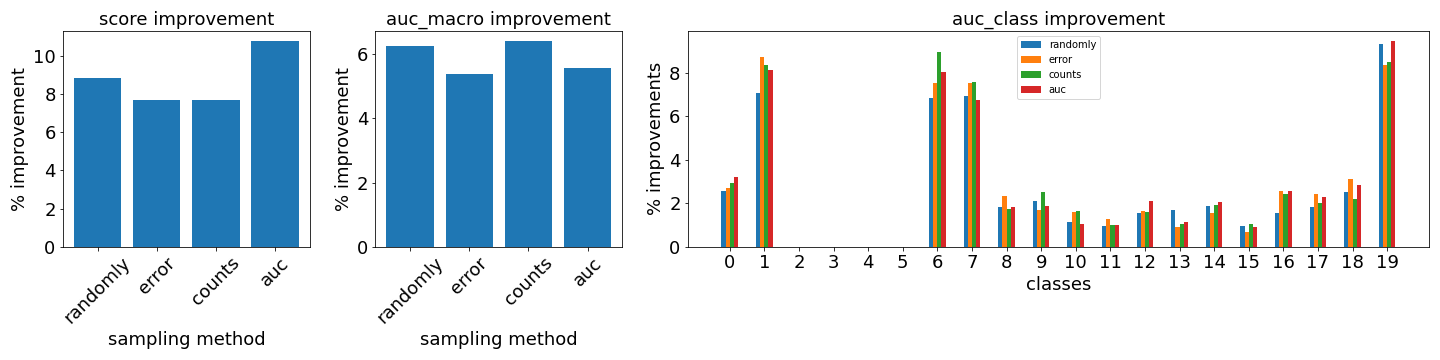"}
    \caption{Performance improvement on different metrics over different sample selection methods discussed in Section \ref{ss} using NNModel. The graphs above use the following notation: score refers to predictive accuracy, auc\_macro refers to macro-AUC score, and auc\_class refers to AUC score per class.}
    \label{fig:ssdistrbution}
\end{figure}

From Figure \ref{fig:ssdistrbution}, we show the max improvements over different sampling methods from Section \ref{ss} over 50 trials. We note that PPSS method, or the AUC approach, demonstrates highest improvements in predictive accuracy, however, based on the macro-AUC and class AUC scores, there doesn't seem to be a clear method that outperforms the others, all performing comparatively targeting the minority classes than the traditional sampling methods seen in \ref{fig:benchmark}.

\begin{figure}
    \centering
    \includegraphics[width =\textwidth]{"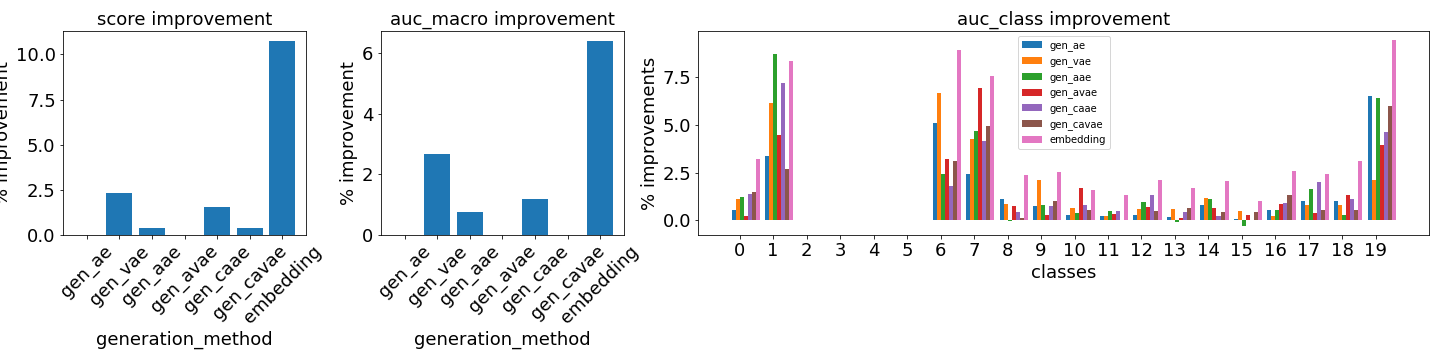"}
    \caption{Performance improvement on different metrics over different generation methods discussed in Section \ref{dgm} using NNModel. The graphs above use the same notation as Figure \ref{fig:ssdistrbution}. In this graph, gen\_ae represents generative auto-encoder, gen\_vae represent generative VAE, gen\_aae represents generative adversarial auto-encoder, gen\_avae represent generative adversarial VAE, gen\_aae represents generative conditional adversarial auto-encoder, and gen\_cavae represent generative conditional adversarial VAE. }
    \label{fig:gmdistrbution}
\end{figure}

We now compare the max improvements over different generative methods, comparing models from Section \ref{dgm} and the traditional embedding-based SDG method in Section \ref{SDG}, over 50 trials. From Figure \ref{fig:gmdistrbution}, we note that DGMs demonstrates very minimal improvements in predictive accuracy and macro-AUC score, and does not perform comparatively to the embedding-based SDG methods. We assume this to be related to the issue of over-fitting, as we accommodated this issue only by employing early stopping with basic heuristics: if the majority of training data ($\geq 50\%$) share $\geq 80 \%$ of the features to the target sample, then we end training. We suggest for future works to handle this issue of over-fitting in the context of SDG more efficiently. 

\begin{figure}
    \centering
    \includegraphics[width =\textwidth]{"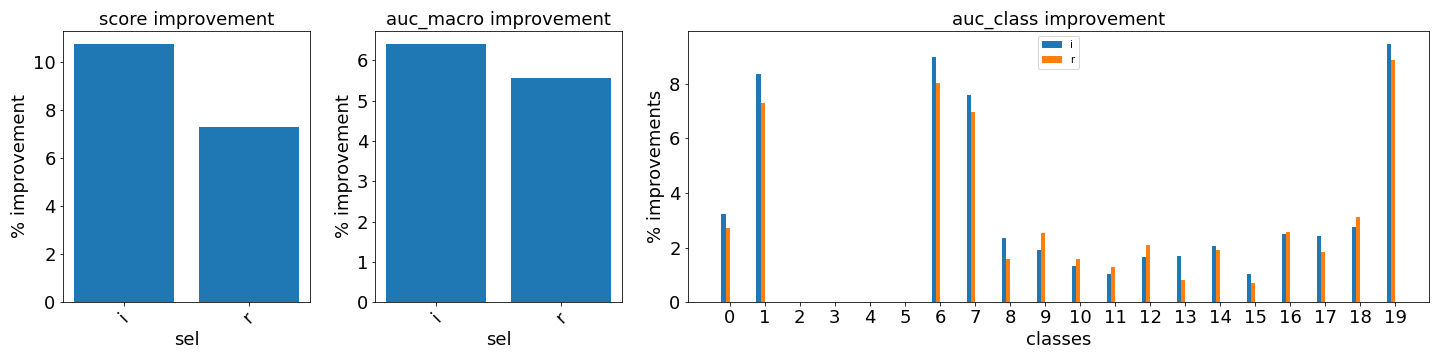"}
    \caption{Performance improvement on different metrics over different feature selection methods discussed in Section \ref{fs} using NNModel. The graphs above use the same notation as Figure \ref{fig:ssdistrbution}. The letter i represents the use of some feature weighing, such as feature importance or feature imbalance, for feature selection, and the letter r represent random feature selection.}
    \label{fig:fdistrbution}
\end{figure}

Similarly, the max improvements over different feature selection methods when using embedding-based SDG, discussed in Section \ref{fs}, over 50 trials. In Figure \ref{fig:fdistrbution}, we confirm that weighing feature selection is more effective than random feature selection in all performance metrics. For both accuracy and macro-AUC, the difference is obvious, and for class AUC scores, we can see that majority of the minority classes, as well as the majority classes, are handled better with weighed feature selection.

\begin{figure}
    \centering
    \includegraphics[width =\textwidth]{"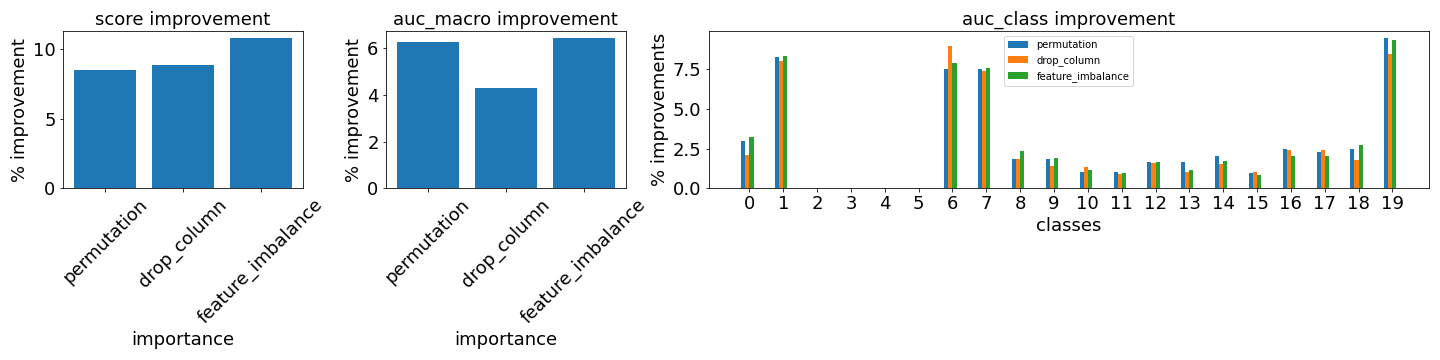"}
    \caption{Performance improvement on different metrics over different feature weighing approaches discussed in Section \ref{fs} using NNModel. The graphs above use the same notation as Figure \ref{fig:ssdistrbution}.}
    \label{fig:fidistrbution}
\end{figure}

Figure \ref{fig:fidistrbution} shows the max improvements over different feature weighing methods used in feature selection, discussed in Section \ref{fs}, over 50 trials. The feature imbalance approach performs better in all scoring metrics, with permutation importance and drop-column importance, the two feature importance methods, performing similarly in predictive accuracy. However, permutation importance performs noticeably better in macro-AUC and class AUC than drop-column importance.

\begin{figure}
    \centering
    \includegraphics[width =\textwidth]{"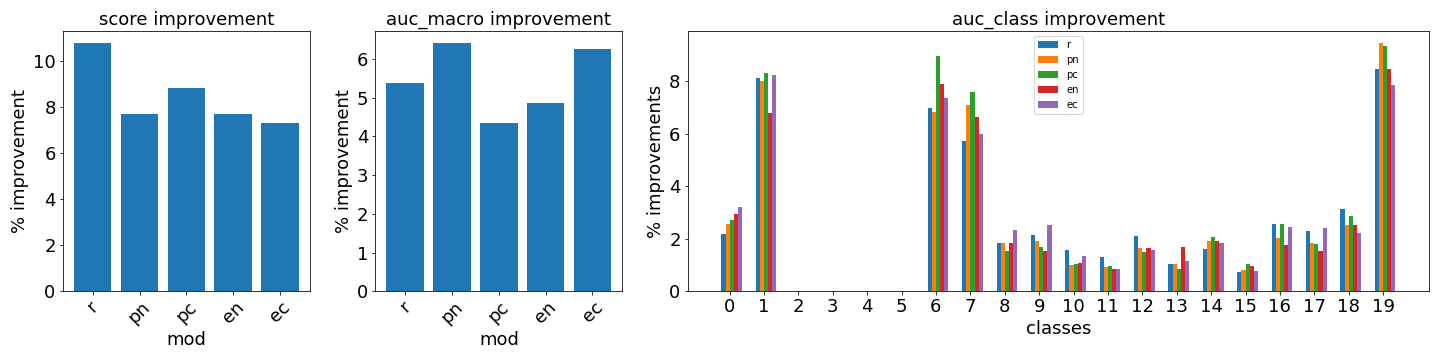"}
    \caption{Performance improvement on different metrics over different feature modification methods discussed in Section \ref{fs} using NNModel. The graphs above use the same notation as Figure \ref{fig:ssdistrbution}. In this graph, r represents random modification, pn represents nearest neighbor search on features reduced by principal component analysis (PCA), pc represents cosine similarity on features reduced by PCA, en represents nearest neighbor search on embedded features, and ec represents cosine similarity on embedded features}
    \label{fig:moddistrbution}
\end{figure}

The max improvements over different feature modification methods used, discussed in Section \ref{fm}, over 50 trials is seen in Figure \ref{fig:moddistrbution}. In terms of macro-AUC score, modification approaches using nearest neighbor search on features reduced by PCA and cosine similarity on embedded features both outperformed other methods. We see that random feature modification does better in terms of predictive accuracy. For class AUC score, we can see that cosine similarity on reduced features by PCA helps the minority classes the most, however this approach does perform worse in terms of macro-AUC score than random modification.

\begin{figure}
    \centering
    \includegraphics[width =\textwidth]{"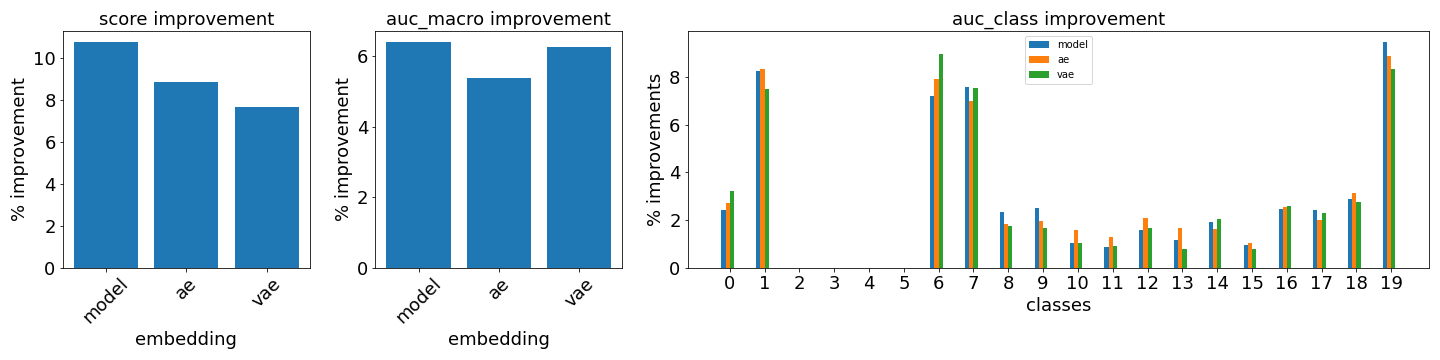"}
    \caption{Performance improvement on different metrics over different embedding methods discussed in Section \ref{fs} using NNModel. The graphs above use the same notation as Figure \ref{fig:ssdistrbution}. In this graph, model represents using the embedding learned by the NNModel for feature modification, ae represents using the embedding learned by an independent auto-encoder for feature modification, and vae represents using the embedding learned by an independent VAE for feature modification.}
    \label{fig:edistrbution}
\end{figure}

We show the max improvements over different embedding generation methods for feature modification, discussed in Section \ref{embgen}, over 50 trials in Figure \ref{fig:edistrbution}. We find that for both accuracy and macro-AUC score, using the embedding matrix from the NNModel outperform the other methods, with VAE being a close second in terms of macro-AUC score. For class AUC scores, we find that VAEs and embedding matrix from NNModel improve the scores of the minority classes the most.

\begin{figure}[]
    \centering
    \includegraphics[width =\textwidth]{"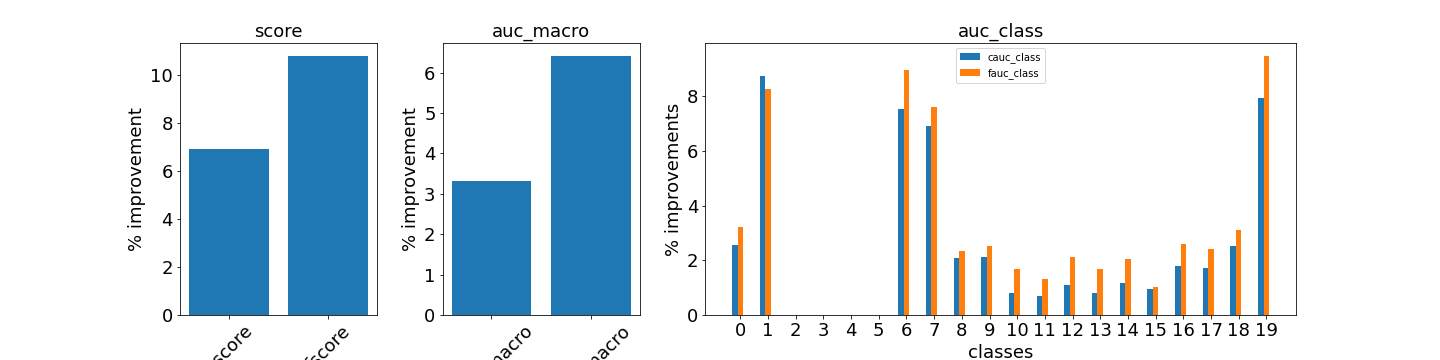"}
    \caption{Performance improvement on different metrics over different synthetic sample usage discussed in Section \ref{ssu} using NNModel. The graphs above use the same notation as Figure \ref{fig:ssdistrbution}. In this graph, any scores denoted with a prefix f means cold start training was used, and score denoted with a prefix c means warm start training was used.}
    \label{fig:ssudistrbution}
\end{figure}

In Figure \ref{fig:ssudistrbution}, we show the max improvements over different synthetic sample usage approaches discussed in Section \ref{ssu} over 50 trials. We find that for all performance metrics, cold start performs better than warm start approaches.

\begin{figure}[]
    \centering
    \includegraphics[width = \textwidth]{"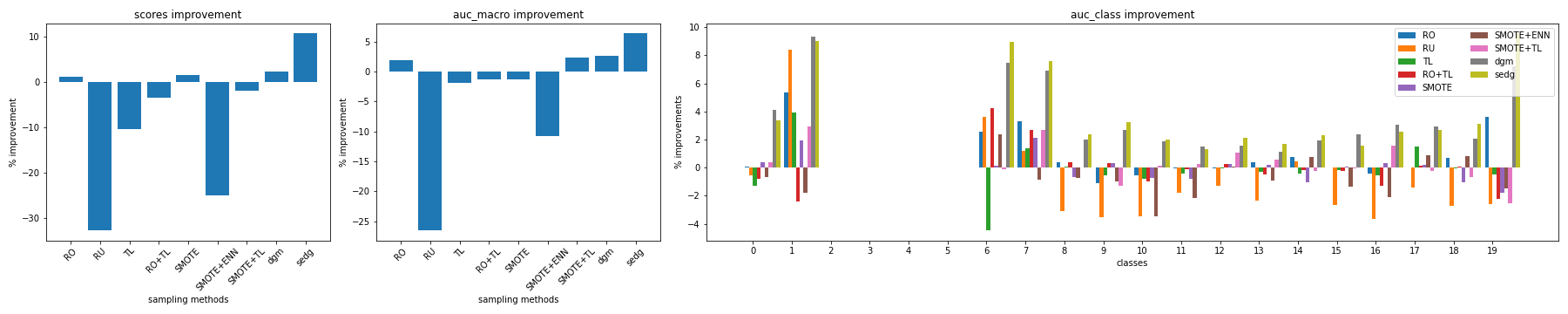"}
    \caption{Performance improvement comparing traditional sampling methods against DGMs and SEDG using NNModel as classifier. The graphs above use the same notation as Figure \ref{fig:ssdistrbution}.}
    \label{fig:finaldldistrbution}
\end{figure}

In Figure \ref{fig:finaldldistrbution}, we find that the SEDG method outperforms all classic resampling methods for all performance metrics when using NNModel as the classifier. In fact, many of the classic sampling method perform poorly, harming the performance, even given 50 trials to run. Most notably, the class-AUC score demonstrate a significant positive difference in targeting the minority classes when using SEDG methods.
 
\subsection{Traditional Classifiers}
Now we consider the traditional classifiers seen in Section \ref{classifiers}, and how SDG methods affect their performances. First, we show that when we transfer the embeddings from the deep models, to act as data pre-processing mappings, to these learners, their performance increases noticeably, as seen in Table \ref{tab:emb_improve}. This supports and allows us to proceed to use embedding-based SDG methods without worries of the embeddings being incompatible to these classifiers.

\begin{table}[]
\centering
\begin{tabular}{c|c}
Model         & \% improvement \\ \hline
OvR SVM       & 8.4615         \\ \hline
OvO SVM       & 3.0769         \\ \hline
Random Forest & 13.846         \\ \hline
XGBoost       & 4.2307         \\ \hline
\end{tabular}
\caption{Performance improvement when using trained NNModel's embedding as data pre-processing to following classifiers. In the table above, OvR SVM represents one-vs-rest SVM and OvO SVM represents one-vs-one SVM.}
\label{tab:emb_improve}
\end{table}

Thus, we test the embedding-based SDG methods and the deep generative models on the traditional classifiers similar to Section \ref{nntest}.

\begin{figure}[]
    \centering
    \includegraphics[width = \textwidth]{"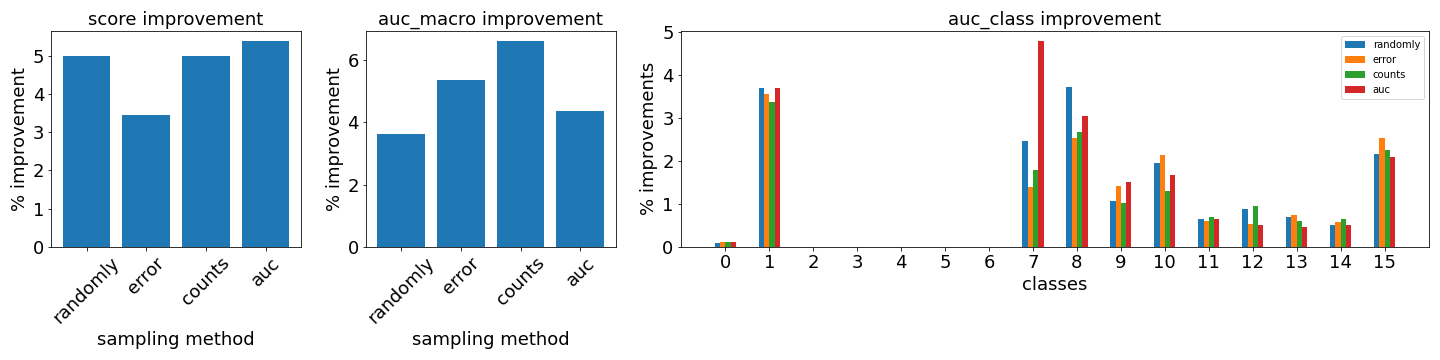"}
    \caption{Performance improvement on different metrics over different sample selection methods discussed in Section \ref{ss} using traditional classifiers. The graphs above use the same notation as Figure \ref{fig:ssdistrbution}.}
    \label{fig:tsm_distribution}
\end{figure}

From Figure \ref{fig:tsm_distribution}, we show the max improvements over different sampling methods from Section \ref{ss} over 50 trials. Unlike to the results from Section \ref{nntest}, PaSS method demonstrates highest improvements in macro-AUC and a close second in accuracy, however, for class AUC score and accuracy, PPSS method performs the best.

\begin{figure}[]
    \centering
    \includegraphics[width = \textwidth]{"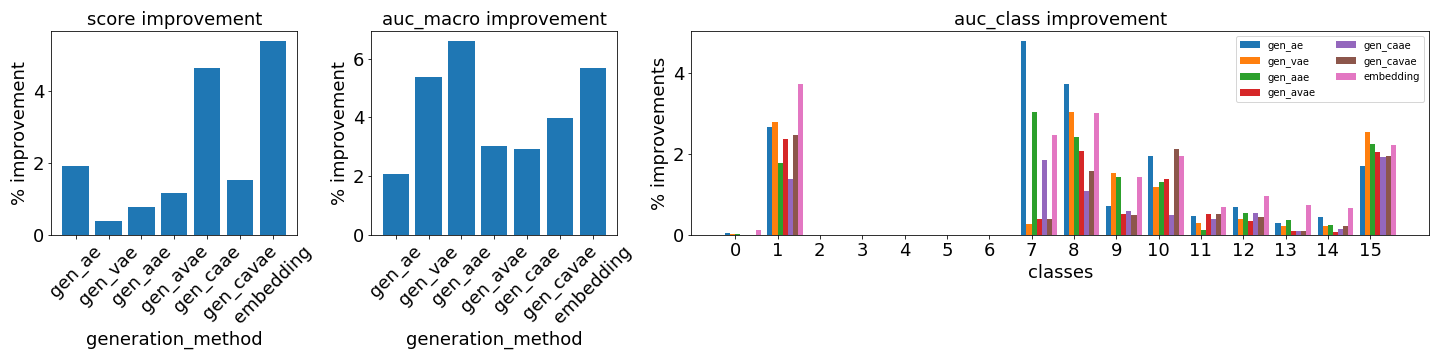"}
    \caption{Performance improvement on different metrics over different generation methods discussed in Section \ref{dgm} using traditional classifiers. The graphs above use the same notation as Figure \ref{fig:gmdistrbution}.}
    \label{fig:tgmdistrbution}
\end{figure}

The max improvements over different generative methods, discussed in Section \ref{dgm}, compared to SEDG methods from Section \ref{SDG} over 50 trials is seen in Figure \ref{fig:tgmdistrbution}. We see that SEDG methods performs the best in accuracy, but generative adversarial autoencoder outperforms SEDG methods in terms of macro-AUC score. For class AUC score, generative autoencoder seem to perform the best in terms of improving the minority classes.

\begin{figure}[]
    \centering
    \includegraphics[width = \textwidth]{"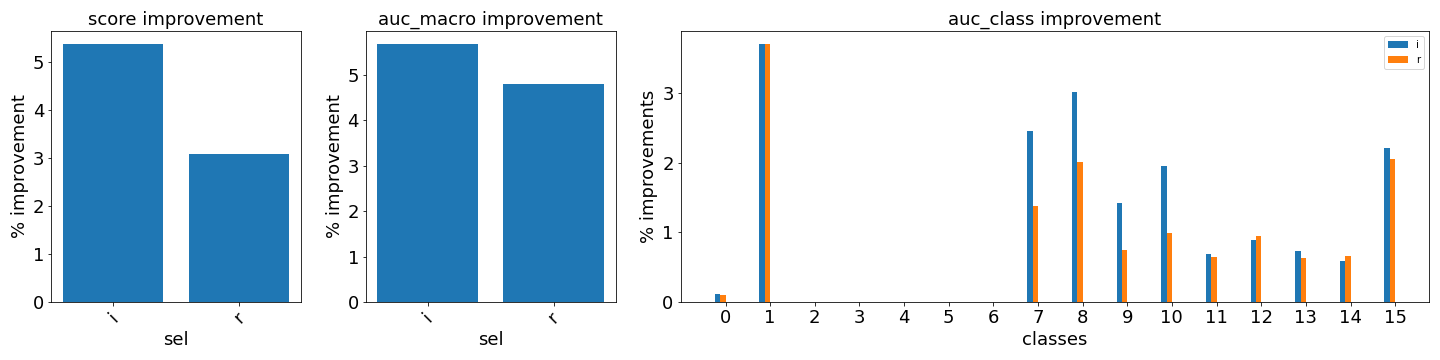"}
    \caption{Performance improvement on different feature selection methods discussed in Section \ref{fs} using traditional classifiers. The graphs above use the same notation as Figure \ref{fig:fdistrbution}.}
    \label{fig:tsdistrbution}
\end{figure}

We find the max improvements over different feature selection methods from Section \ref{fs} used in embedding-based SDG method in Section \ref{SDG}, over 50 trials to again confirm that weighted feature selection proves to be more effective than random feature selection, seen in Figure \ref{fig:tsdistrbution}.

\begin{figure}[]
    \centering
    \includegraphics[width = \textwidth]{"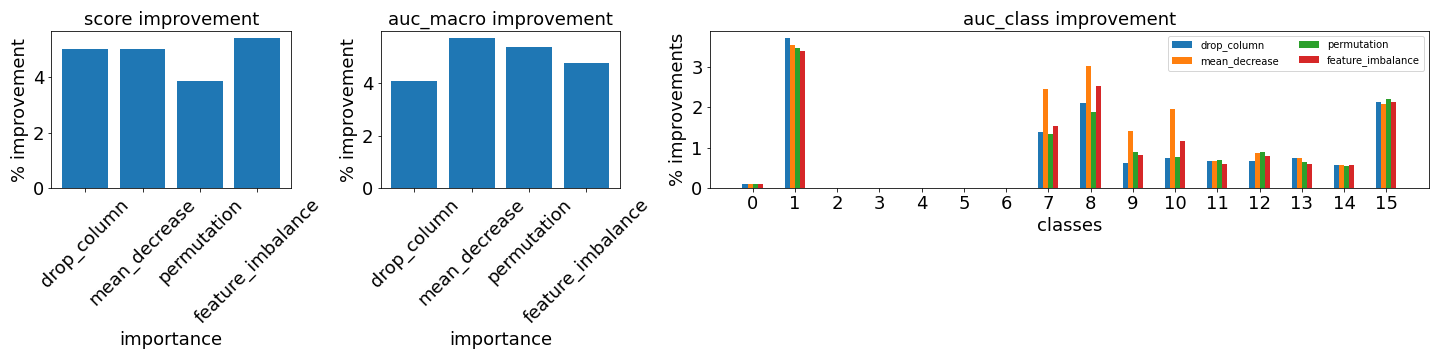"}
    \caption{Performance improvement on different feature weighing methods discussed in Section \ref{fi} using traditional classifiers. The graphs above use the same notation as Figure \ref{fig:fidistrbution}}
    \label{fig:tfidistrbution}
\end{figure}

Figure \ref{fig:tfidistrbution} shows the max improvements over different feature weighing methods used in feature selection, discussed in Section \ref{fs}, over 50 trials. Feature imbalance again performs best in terms of accuracy, but falters in other performance metrics. For macro-AUC and class AUC scores, mean decrease importance for tree-based classifiers seems to improve the minority classes the most than the other approaches.

\begin{figure}[]
    \centering
    \includegraphics[width = \textwidth]{"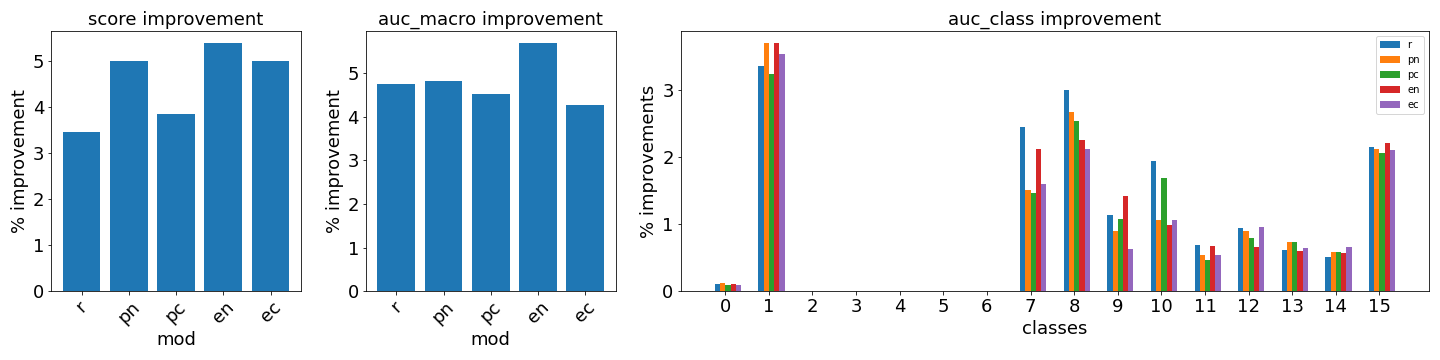"}
    \caption{Performance improvement on different metrics over different feature modification methods discussed in Section \ref{fs} using traditional classifiers. The graphs above use the same notation as Figure \ref{fig:moddistrbution}}
    \label{fig:tmoddistrbution}
\end{figure}

The max improvements over different feature modification methods used, discussed in Section \ref{fm}, over 50 trials is seen in Figure \ref{fig:tmoddistrbution}. All modification methods performed similarly in terms of class AUC scores, with random modification taking a slight lead. In terms of accuracy and macro-AUC, nearest neighbor on embedded features  was the most successful.

\begin{figure}[]
    \centering
    \includegraphics[width = \textwidth]{"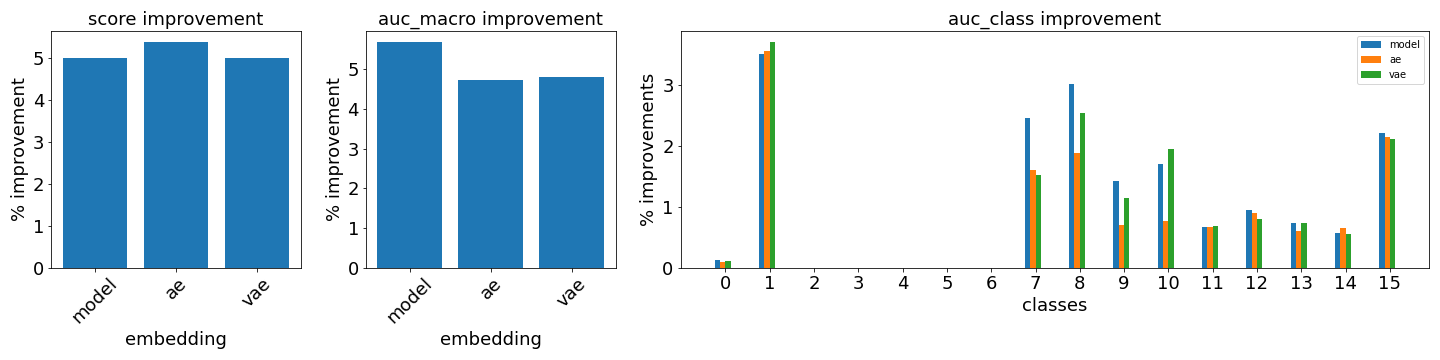"}
    \caption{Performance improvement on different embedding methods discussed in Section \ref{embgen} using traditional classifiers. The graphs above use the same notation as Figure \ref{fig:ssdistrbution}.}
    \label{fig:tedistrbution}
\end{figure}

We show the max improvements over different embedding generation methods for feature modification, discussed in Section \ref{embgen}, over 50 trials in Figure \ref{fig:tedistrbution}. We find that for accuracy, autoencoder outperform the other methods, with VAE and embedding matrix from NNModel performing similarly. For both macro-AUC and class AUC score, using the embedding matrix from NNModel proves to be the best.

\begin{figure}[]
    \centering
    \includegraphics[width = \textwidth]{"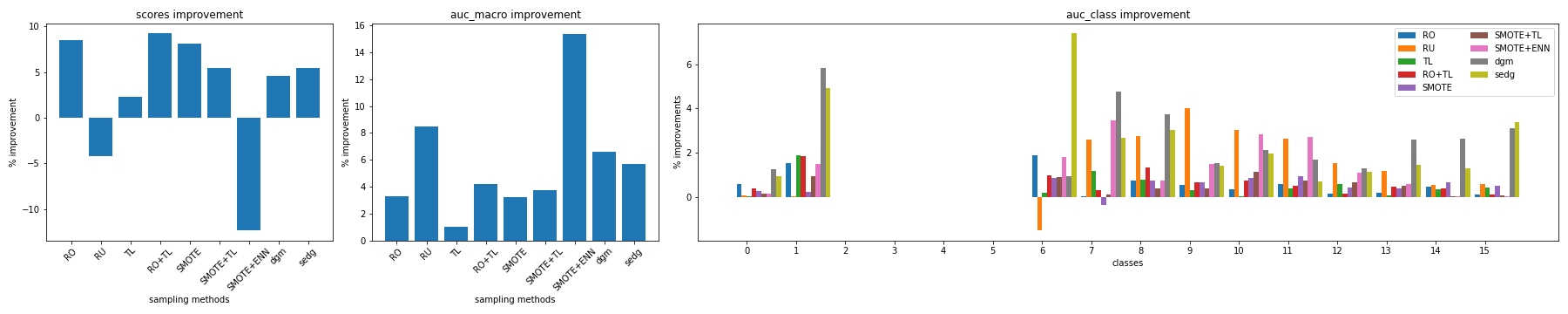"}
    \caption{Performance improvement comparing classic re-sampling methods against DGMs and SEDG with traditional machine learning classifiers. The graphs above use the same notation as Figure \ref{fig:ssdistrbution}.}
    \label{fig:finaltmldistrbution}
\end{figure}

In Figure \ref{fig:finaltmldistrbution}, we find that the SEDG method, while it does not excel in accuracy and macro-AUC score \% improvements for traditional machine learning classifiers, it is able to target the minority classes more effectively in class-AUC score than all other methods. Additionally, compared to the other classic re-sampling methods, the results show when considering all the performance metric, SEDG methods, similarly to DGMs, performs the best as certain re-sampling approaches that excel in one metric often fail to replicate the same success in other metrics.

\begin{figure}[]
    \centering
    \includegraphics[width = \textwidth]{"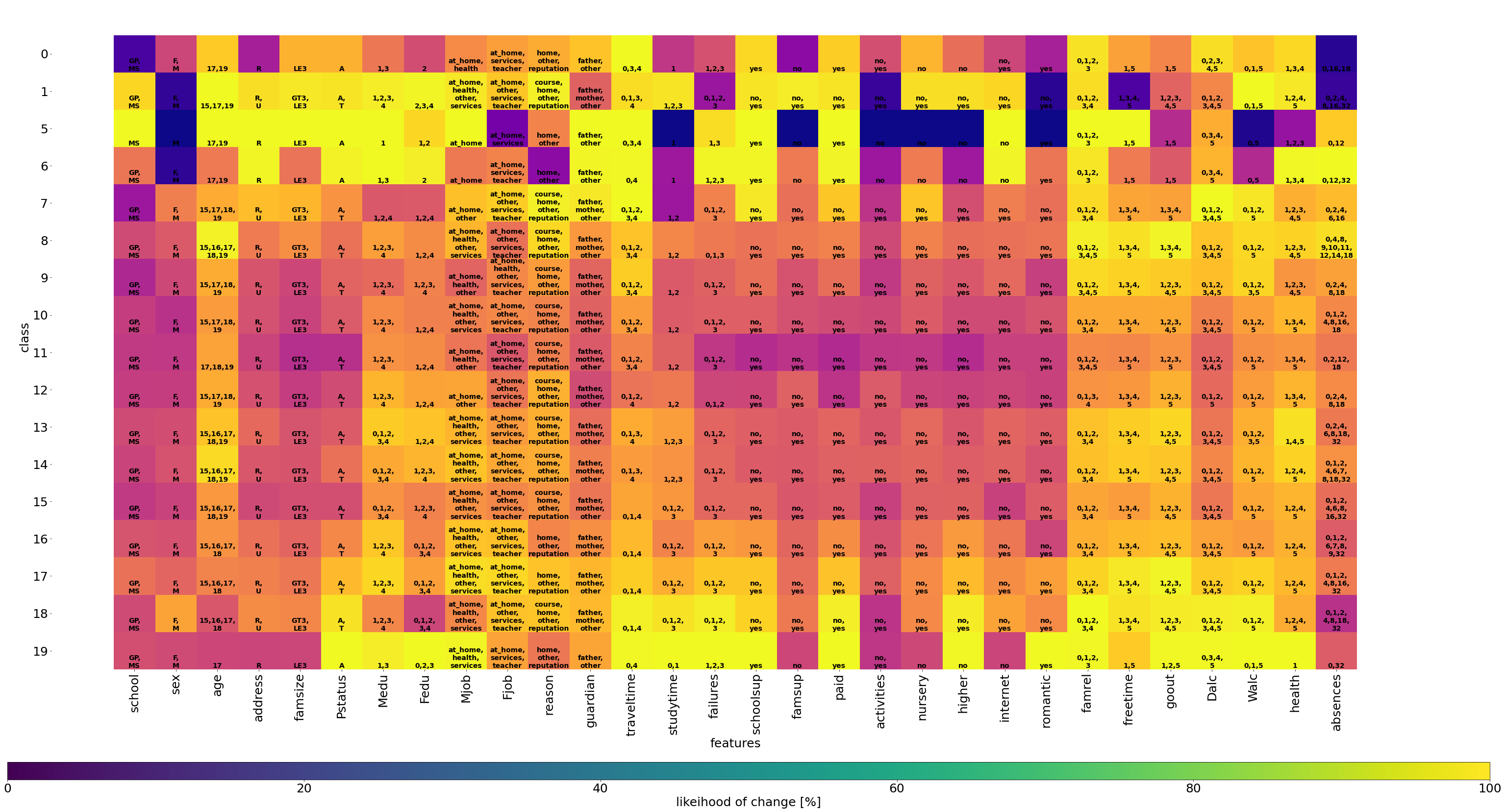"}
    \caption{Likelihood of feature modification and feature value candidates using generative VAE from 100 synthetic samples per class.}
    \label{fig:vae}
\end{figure}

\subsection{Understanding Student Performance}
In this section, we investigate the synthetic samples more in-depth for each class to improve our understanding for student performance by showing how likely certain features are allowed to change without much significant changes to the data distribution, similar to feature importance. However, with synthetic samples, we can also see how we can change these features, thus seeing possible feature value candidates.

In Figure \ref{fig:vae}, we use a VAE as DGM to create $100$ synthetic samples for each class to show how likely each feature in each class is subject to change, and what feature values is reasonable. For example, we can see for low performing students, the likelihood of changing absences is low, with greater number of absences being a possible candidates. We also see that some features, such as travel time, school supplies, and famrel, are highly susceptible to change, inferring to their lack of distinction between possible candidates.

\section{Conclusion}
In this paper, we proposed methods for embedding-based SDG and investigated DGMs for class imbalanced tasks, specifically student performance tasks. We tested SEDG approach and DGMs against standard balancing methods, and found greater improvements in our proposed approaches in terms of \% improvement in accuracy, macro-AUC score, and class-AUC score when we use our NNModel as the classifier, and an more comprehensive improvement when we use traditional machine learning classifiers on student performance task. We also introduced a technique for greater interpretibilty for SDG methods by looking at the synthetic samples generated and seeing how likely each feature is modified and to what values it can take on for each class.

\section{Acknowledgement}
We thank Samuel Schmidgall (George Mason University), Ji Kuo (George Mason University), and Jay Deorukhkar (George Mason University) for helping create and build up the core ideas of the SEDG framework, and their efforts in paper revisions. We thank Dr. Huzefa Rangwala (George Mason University) for useful suggestions and advices that motivated key components such as weighted feature selection. We thank Yuanqi Du (George Mason University) for assistance in selecting the area of focus to be the educational domain and in finding the appropriate dataset.

\bibliography{reference}{}
\bibliographystyle{plain}

\end{document}